\documentclass[letterpaper, 10 pt, conference]{ieeeconf}  

\IEEEoverridecommandlockouts                              

\title{\LARGE \bf
Efficient and Interaction-Aware Trajectory Planning for Autonomous Vehicles with Particle Swarm Optimization
}

\author{Lin Song$^{2}$  \hspace{.5 in} David Isele$^{1}$ \hspace{.5 in} Naira Hovakimyan$^{2}$  \hspace{.5 in} Sangjae Bae$^{1}$
\thanks{*This work was supported by Honda Research Institute, USA. }
\thanks{$^{1}$Honda Research Institute, San Jose, CA 95134, USA.
        \{\tt\small disele, sbae\}@honda-ri.com}%
\thanks{$^{2}$ University of Illinois Urbana-Champaign, Urbana, IL 61801, USA.
        \{\tt\small linsong2, nhovakim\}@illinois.edu}%
}

\usepackage{color,soul}
\usepackage{xcolor}
\usepackage{graphicx}
\usepackage{amsmath}
\usepackage{algorithm}
 \usepackage{algpseudocode}
 \usepackage{algorithmicx}
 \algdef{SE}[DOWHILE]{Do}{doWhile}{\algorithmicdo}[1]{\algorithmicwhile\ #1}%
\newtheorem{remark}{Remark}
\algrenewcommand\algorithmicrequire{\textbf{Input:}}
\algrenewcommand\algorithmicensure{\textbf{Output:}}

\usepackage{cleveref}

\usepackage{cite}

\usepackage{subcaption}
\usepackage{mwe}
\begin{document}
\maketitle
\thispagestyle{empty}
\pagestyle{empty}

\begin{abstract}This paper introduces a novel numerical approach to achieving smooth lane-change trajectories in autonomous driving scenarios. 
Our trajectory generation approach leverages particle swarm optimization (PSO) techniques, incorporating Neural Network (NN) predictions for trajectory refinement.
The generation of smooth and dynamically feasible trajectories for the lane change maneuver is facilitated by combining polynomial curve fitting with particle propagation, which can account for vehicle dynamics. The proposed planning algorithm is capable of determining feasible trajectories with real-time computation capability. We conduct comparative analyses with two baseline methods for lane changing, involving analytic solutions and heuristic techniques in numerical simulations. The simulation results validate the efficacy and effectiveness of our proposed approach. 

\end{abstract}

\section{INTRODUCTION}



\begin{figure}
\includegraphics[width=8cm, height=3.5cm]{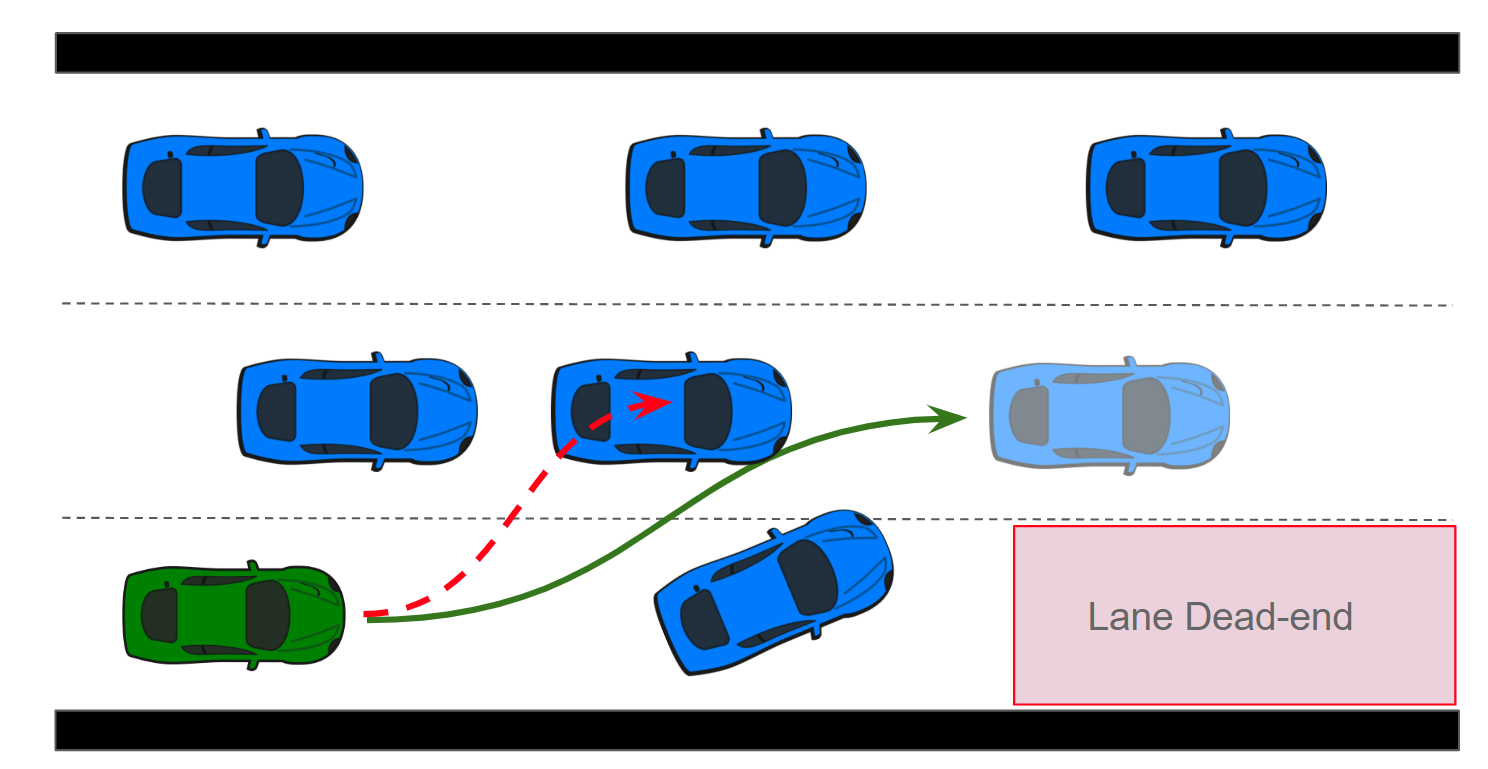}
\centering
\caption{Motivation example: lane change in dense traffic.}\label{fig:motivation}
\end{figure}

We consider motion planning for autonomous vehicles in highly dense traffic scenarios, as depicted in Figure~\ref{fig:motivation}. 
The complex driving scenario necessitates the development of decision-making strategies that consider interactions and aim to mitigate excessive caution during the merging maneuvers. The complexity of the potential interactions has motivated the development of game-theoretic approaches~\cite{sadigh2018planning,schwarting2019social,isele2019interactive,tian2021anytime}, but these are often computationally expensive and do not scale well to dense traffic settings.

Reinforcement learning (RL) techniques have been effectively employed to address the intricate task of interaction-aware trajectory planning. These techniques encompass both explicit learning of control policies and implicit learning of cost functions. A recognized study is on the utilization of inverse reinforcement learning (IRL) to learn human driver rewards, subsequently planning more efficient behavior for autonomous vehicles and accomplishing interaction-aware planning~\cite{sadigh2016planning}. Lee et al.~\cite{lee2022spatiotemporal} further leverage IRL to learn costmaps for model predictive controllers (MPCs) based on human driver demonstrations. A model-free RL agent is constructed for lane-change control within dense traffic settings, showing its capability of interacting and opening a gap to accomplish the merge maneuver~\cite{saxena2020driving}. While efforts have been made to shrink the gap concerning reliability and interpretability in deploying RL techniques in practice, such as safe RL~\cite{isele2018safe,shalev2016safe} and adversarial RL~\cite{gupta2022towards}, the challenge persists in achieving this integration.



Considering that feed-forward evaluations in already-trained neural networks can achieve real-time efficiency, an alternative approach is to use neural networks (NN) to handle the interactive component. One example of this is
Neural Network Model Predictive Control (NNMPC)~\cite{bae2022lane, bae2020cooperation}, which attempts to solve merging in dense traffic by combining data-driven interactive prediction, such as social generative adversarial networks (SGANs)~\cite{gupta2018social}, with control-theoretic mechanisms, namely model predictive control (MPC). This integration is intended for producing smooth and interpretable trajectories while preserving the powerful estimation engines of learning-based strategies. 

In NNMPC, trajectory candidates can be either computed from random samples~\cite{bae2020cooperation} or generated as spiral curves contingent upon driving intentions~\cite{bae2022lane}. However, sampling-based strategies introduce a potential accuracy-efficiency trade-off where a large number of trajectories must be generated to find a near-optimal solution. Meanwhile, the non-convexity of the problem makes it difficult to optimize efficiently; for example, obtaining an analytic solution for MPC with a neural network component employs the Alternating Direction Method of Multiplier (ADMM) technique in~\cite{gupta2023interaction}, which finds a provably optimal solution, but at a runtime that is not practical for real-time applications. In this work, we investigate the use of particle swarm optimization (PSO) to improve the accuracy-efficiency tradeoff in the trajectory generation step. 

The particle swarm optimization (PSO) algorithm employs an intelligent mechanism facilitating collaborative search and knowledge sharing among its population~\cite{gad2022particle}, establishing its efficacy in addressing complex non-linear optimization problems. Notably, PSO has demonstrated its capacity for fast convergence with a few parameter adjustments required~\cite{pal2013robot}. While the standard PSO formulation excels in solving unconstrained optimization problems, constraints can be integrated into its cost function (alternatively a fitness function). An accelerated variant of PSO introduced in~\cite{arrigoni2019non} simplifies particle velocity updates for improved performance.

In robotics, applications of PSO in solving optimization problems involve bipedal robot footstep placement in real-time~\cite{hong2019real}, path planning with imprecise environment measurements~\cite{zhang2013robot}, and multi-robot coordination~\cite{thabit2018multi}.
In~\cite{biswas2017obstacle} and~\cite{roberge2012comparison}, position coordinates are interpreted as particles guiding the derivation of safe, optimal, and feasible paths within the PSO algorithm, albeit neglecting model dynamics. Furthermore, Kim et al.~\cite{kim2022hierarchical} utilized PSO for short-term planning, through encoding steering angles as particles to generate feasible trajectories within a hierarchical motion planning framework.  

While previous endeavors have explored the utilization of PSO formulations in trajectory planning for autonomous systems, there remains a significant gap in deploying numerical PSO techniques for interaction-aware trajectory planning when other vehicles' behavior is characterized by neural networks.
In this work, our main contribution lies in integrating PSO and polynomial curve fitting approaches with a learning-based prediction module, aiming to ensure safety, trajectory smoothness, and real-time applicability in interaction-aware planning. In the context of lane changes, neural networks are utilized to model the behaviors of nearby vehicles; PSO is employed to solve the associated constrained optimization problem, and a polynomial curve fitting procedure ensures the smoothness of the generated trajectory. The proposed algorithm computes dynamically feasible trajectories that leverage interactive predictions and achieve interaction-aware lane changes in a real-time manner.

\section{SYSTEM AND MODEL}
\subsection{Vehicle Model}
We utilize the nonlinear kinematics bicycle model in~\cite{kong2015kinematic} for trajectory planning. The discrete-time kinematics are represented as:
\begin{subequations}
\label{eq:veh-kinematics}
    \begin{align}
    x_{k} &= \Delta t \cdot v_{k-1}\cos(\psi_{k-1}+\beta_{k-1}) + x_{k-1}, \label{eq:kin-xpos}\\
    y_{k} &= \Delta t \cdot v_{k-1} \sin(\psi_{k-1}+\beta_{k-1}) + y_{k-1}, \label{eq:kin-ypos}\\
    \psi_{k} &= \frac{v_{k-1}}{l_r}\sin(\beta_{k-1}) + \psi_{k-1}, \label{eq:kin-heading}\\
    v_k &= \Delta t \cdot a_{k-1} + v_{k-1}, \label{eq:kin-vel} \\
    \beta_k &= \tan^{-1}(\frac{l_r}{l_f+l_r}\tan{(\delta_{k})}),\label{eq:kin-beta}
\end{align}
\end{subequations}
where $(x_k,y_k)$ represents the Cartesian coordinate of the center of the vehicle at time $k$, $\psi_k$ denotes the inertial heading, $v_k$ is the vehicle speed, $a_k$ is the acceleration of vehicle center, $l_f$ and $l_r$ denote the distance from the center of the vehicle to the front and rear axles, respectively; $\beta_k$ is the angle of the current velocity of the center to the longitudinal axis of the vehicle. The control inputs of the bicycle model include steering angle $\delta_k$ and acceleration $a_k$. The simple bicycle kinematics model is capable of preserving accuracy while improving computation efficiency. 
\subsection{Prediction Model} \label{sec:SGAN-intro}
Given that the inter-vehicle distances cannot be measured in the trajectory planning phase, we utilize the social generative adversarial network (SGAN)
model~\cite{gupta2018social} to predict vehicular behaviors, thereby facilitating interaction-aware trajectory planning. A trained SGAN model is capable of efficiently generating the most probable trajectories for surrounding vehicles using their positional observations as its inputs. Conceptually, SGAN can be interpreted as a function $\phi(\cdot)$, translating past observations to anticipated positional sequences, represented as $\phi(\cdot): Z(k) \to \hat{Z}(k)$, where

{\small
\begin{equation*}
    Z(k)= \begin{bmatrix}
    z_1(k) & \cdots & z_{N_{veh}}(k)\\
        \vdots & \ddots & \vdots \\
        z_1(k-N_{obs}+1) & \cdots & z_{N_{veh}}(k-N_{obs}+1) 
    \end{bmatrix}\end{equation*}}
 and 
 {\small\begin{equation*}
    \hat{Z}(k) = \begin{bmatrix}
        \hat{z}_1(k+1) & \cdots & \hat{z}_{N_{veh}}(k+1)\\
        \vdots & \ddots & \vdots \\
        \hat{z}_1(k+N_{pred}) & \cdots & \hat{z}_{N_{veh}}(k+N_{pred})
    \end{bmatrix}.\end{equation*}}
Here, $z_i(k)= (x_i(k), y_i(k))$ denotes a tuple of position coordinates for time $k$, $N_{obs}$ and $N_{pred}$ represent the observation and prediction horizons, respectively, and $N_{veh}$ represents the number of total vehicles. 

In this work, the SGAN model we use is trained on simulation data. The training and validation datasets are collected from multiple driving scenarios with various traffic densities from a defined driver model. The longitudinal dynamics of the driver model are controlled by an intelligent driver model (IDM)~\cite{treiber2000congested}, and the lane-changing behavior is governed by the strategy of Minimizing Overall Braking Induced by Lane changes (MOBIL)~\cite{kesting2007general}.  Interested readers can refer to~\cite{bae2022lane} for more details about the SGAN training.

\subsection{Particle Swarm Optimization (PSO)}\label{sec:pso-prelim}
PSO algorithm imitates the social behavior of collaborative search and information exchange within swarms and was initially proposed in~\cite{kennedy1995particle}. Due to its derivative-free nature, the PSO algorithm is well-known for solving nonlinear optimization problems with real-time computation capability. 

In PSO, each particle in the multi-dimensional search space represents a candidate solution to the optimization problem; these particles also memorize the best performance from their past search. The positions $p_i$ and velocities $\nu_i$ of the particles are randomly initialized according to a uniform distribution for exploration purposes.   We assign the range for the uniform distribution of initial particle position $p_i$ and velocity $v_i$ based on information computed from an initial trajectory plan according to an existing planner. This approach enhances the efficiency of feasible solution exploration. The particle velocity $\nu_i$ is then adjusted according to the best self-cognitive experience $p_{i,lb}$ and the best experience achieved by the entire population $p_{gb}$, as well as the velocity component to be preserved. The particle velocity update rule of the standard PSO algorithm is 
\begin{equation}\label{eq:pso-conv-vel}
    \nu_{i,k+1} = w\nu_{i,k} + c_1r_1 (p_{i,lb} - p_{i,k}) + c_2r_2  (p_{gb} - p_{i,k}),
\end{equation}
where $w$ is the inertia weight, $c_1, c_2$ are acceleration coefficients reflecting self-cognition and social influence, and random numbers $r_1, r_2 \in [0,1]$ are deployed to avoid local optima in the optimization process. The random initialization of particle positions, velocities, and the stochastic adjustment of particle velocities, facilitate the efficient search for feasible solutions. Large inertia weight $w$ favors the global search of an optimal solution, while small inertia weight $w$ improves the local search capacity. A dynamically reduced inertia weight is suggested in~\cite{shi1998parameter} for faster convergence.  The global-best particles are replaced by the local-best particles when a lower cost value is achieved from any agent within the swarm. The updated particle position is then computed using the updated particle velocity according to 
\begin{equation}\label{eq:particle-pos-update}
    p_{i,k} = p_{i,k-1} + \nu_{i,k}.
\end{equation}
\section{APPROACH}
\subsection{Trajectory Planning}
For the task of lane changing in dense traffic conditions, we aim to compute smooth trajectories that enable merging into the target lane without colliding with other vehicles. The problem can be decomposed into two sub-problems: 1) the generation of smooth trajectories, and 2) an optimization scheme taking smooth reference trajectories as inputs while addressing safety constraints posed by surrounding vehicles. 

In addressing the first problem, we utilize polynomial curve fitting techniques. For the latter, we represent vehicle steering angles as particles, and PSO is employed to derive a refined trajectory from the given references. We then enforce trajectory smoothness by fitting a polynomial curve that passes through selected waypoints from the PSO-produced trajectory. Through an iterative procedure solving these two interconnected problems, we establish the optimal trajectory ensuring safety while enabling merging into the target lane.

\subsection{Particle Swarm Optimization (PSO) and Cost Function} 
We next introduce how PSO is applied to determine refined steering angle in path planning for lane change behavior.

\subsubsection{Swarm Formulation}  
 In our approach, each particle $p_i$ represents a sequence of steering angles, defined as $p_i = [\delta_{i,1}, \delta_{i,2}, \dots, \delta_{i, N}]$, where $N$ denotes the horizon. These particles undergo propagation via vehicle kinematics~\eqref{eq:veh-kinematics} to maintain the solution feasibility.
 The inputs of the PSO algorithm consist of reference waypoints in the form of position and velocity profiles (i.e., $(x_i,y_i,v_i)$ tuples) from standard path planners, along with observations of other vehicles' positions $[(z_{i,1},\dots,z_{i,N_{veh}})]_{i=1:N_{obs}}$, where $z_i = (x_i,y_i)$ represents a vehicle's position coordinates in the Cartesian space.
 Here, $N_{veh}$ denotes the number of vehicles and $N_{obs}$ indicates the observation horizon. 
 
 Algorithm~\ref{alg:1} provides an architecture of the PSO algorithm. We start with random initialization of particles' positions and velocities, where each particle represents a steering angle sequence for the lane-change maneuver. In each iteration, positions and velocities are updated individually as per~\eqref{eq:particle-pos-update} and~\eqref{eq:pso-conv-vel}. Each particle retains its optimal solution discovered across iterations, which is referred to as the `local-best' solution. Moreover, the population's best solution (referred to as the `global-best' solution) is updated when a superior performance is achieved. Particle velocity are updated based on its prior velocity, its local-best position, and the population's global-best position. 

\begin{algorithm}    
    \caption{PSO for lane changes}\label{alg:1}
    \begin{algorithmic}[1]
        \Require{Reference waypoints for the ego vehicle $[(\bar{x}_i,\bar{y}_i,\bar{v}_i)]_{i=1:N}$, reference heading angle sequence $[\bar{\psi}_i]_{i=1:N}$, observation of other vehicles' position $[(z_{i,1},\dots, z_{i,N_{veh}})]_{i=1:N_{obs}}$, maximum iteration number of PSO $I_p$, number of candidate particles $S_p$}
        \Ensure{ Refined steering angle sequence $[\delta_1^*, \dots, \delta_N^*]$ and dynamically feasible trajectory $[(\hat{x}_i, \hat{y}_i)]_{i=1:N}$ }
        \For{each particle $i=1,2,\dots,S_p$}
        \State Randomly initialize steering angle sequence $p_i = [\delta_{i,1}, \dots, \delta_{i,N}]$ and its velocity $\nu_i$ according to uniform distribution
        \State Initialize the particle's best-known position to the initial position: $p_{i,lb} \leftarrow p_i$ 
        \EndFor
        \State Initialize the swarm's best position $p_{gb}$
        \While{$iter < I_p$}
        \For {each particle $i=1,2,\dots,S_p$}
         \State Update particle velocity $\nu_i$ according to~\eqref{eq:pso-conv-vel}
         \State Update particle position $p_i$ according to~\eqref{eq:particle-pos-update}
         \State Propagate updated steering angle sequence $p_i$ through kinematics~\eqref{eq:veh-kinematics} and record dynamically feasible trajectory $[(\hat{x}_i,\hat{y}_i)]_{i=1:N}$
         \State Evaluate cost value of the new particle $p_i$ using~\eqref{eq:fitness-all}
         \If {$f(p_i) < f(p_{i,lb})$}
         {$p_{i,lb} \leftarrow p_{i}$}
         \If {$f(p_{i,lb}) < f(p_{gb})$}
             {$p_{gb} \leftarrow p_{i,lb}$}
         \EndIf
         \EndIf
         \EndFor
        \EndWhile
    \end{algorithmic}
\end{algorithm}

\subsubsection{Cost Function for Evaluation}
 Each particle represents a sequence of steering angles and the cost evaluation of particles proceeds in two steps: Initially, the steering angle sequence is propagated through the vehicle dynamics as per~\eqref{eq:veh-kinematics}. This yields dynamically feasible trajectories, ensuring the generated position sequences adhere to all non-holonomic constraints. Subsequently, the cost function evaluates the generated trajectory corresponding to different particles. Given that the conventional PSO formulation is tailored for unconstrained optimization problems, we incorporate safety constraints into the cost function. Considering safety is the primary concern, particle candidates that cannot drive the vehicle to accomplish the lane change safely attain a large cost and will not be favored in the particle search. 
 
Initially, we extract reference waypoints $(x_i,y_i,v_i)$ from conventional planners. Subsequently, we compute the associated reference control sequences -- namely, acceleration $a_i$ and steering angle $\delta_i$ based on the waypoint references. These reference control sequences are utilized in the kinematics propagation to generate dynamically feasible trajectories and in the evaluation of particle costs. The PSO objectives consist of safety considerations, acceleration and jerk regulation for enhanced driving comfort, adherence to the reference trajectory, and alignment with the lane center. For clarity, we present the comprehensive formulation of the cost function $f_{PSO}$ used in PSO:
\begin{equation}\label{eq:fitness-all}
     f_{PSO} = f_{ref} + f_{head} + f_{col} +  f_{a} + f_{j} + f_{s}  + f_{la}.
 \end{equation}
 \normalsize
 The individual components of the function $f_{PSO}$ represent: 
 \begin{itemize}
     \item Reference trajectory deviation: $f_{ref}$ penalizes deviations from the reference trajectory in the form of 
         $f_{ref} = w_{ref}(\sum_{i=1}^{N} (x_i - \bar{x}_{i})^2 + (y_i - \bar{y}_{i})^2)$,
       where $(x_i,y_i)$ are the vehicle's Cartesian coordinates and $(\bar{x}_i,\bar{y}_i)$ refer to the reference trajectory;
     \item Heading deviation: $f_{head}$ penalizes the differences between the vehicle's actual heading $\psi_i$ from the kinematics propagation in~\eqref{eq:kin-heading} and the reference heading $\bar{\psi}_i$ in the form of
$f_{head} = w_{head}(\sum_{i=1}^N(\psi_{i} - \bar{\psi}_{i}))$;
  \item Safety violation: $f_{col}$ applies a significant penalty if certain safety metrics are violated (Details are elaborated in Section~\ref{sec:safety_constraints});
  \item Driving comfort metrics:  $f_{a}, f_{j}$, and $ f_{s}$ relate to driving comfort, capturing smoothness of acceleration and jerk, and steering effort, respectively. $f_a$ and $f_j$ are computed using finite difference over adjacent state trajectory  as follows:
 $
    f_a = 
     w_a[\sum_{i=1}^{N-2} (x_{i+2} - 2x_{i+1} 
    + x_{i}/\Delta t^2)^2 
    + (y_{i+2} - 2y_{i+1} + y_{i}/\Delta t^2)^2]$
and
$f_j = w_j[\sum_{i=1}^{N-3} (-x_{i+3} + 3x_{i+2} -3x_{i+1}+x_{i}/\Delta t^3)^2 
    + (-y_{i+3} + 3y_{i+2} -3y_{i+1}+y_{i}/\Delta t^3)^2]$.
  Furthermore, the $f_s$ term is designed to regulate the steering effort in the form of
$
    f_s = w_s\sum_{i=1}^N |\delta_i|^2$;
\item Lane-center alignment: The term $f_{la}$ is designed to ensure proper alignment with the lane center as the horizon concludes. It is represented as $f_{la}:= w_{la}|y_{N} - y_c|$, where $y_c$ represents the vertical coordinate of the target lane center. Additionally, any particle generates a state trajectory that violates the target lane's boundaries also incurs a penalty. 
 \end{itemize}

 \subsubsection{Collision Avoidance in PSO} In our efforts to avoid moving vehicles during merging, we implement two strategies to enhance both safety and efficiency. The first strategy involves assigning a high-cost value to any steering angle solution that could lead to a collision, which is represented by the  $f_{col}$ term in~\eqref{eq:fitness-all}. This approach ensures such a particle remains a less favorable choice in comparison to other particle candidates. Furthermore, when steering angle solutions with high collision risks are investigated, we also elevate the particle velocity by multiplying the original velocity update~\eqref{eq:pso-conv-vel} with an adaptive coefficient, thereby improving our exploration capacities. 
\subsection{Safety Constraints with SGAN Prediction}\label{sec:safety_constraints}

\subsubsection{Metrics for Inter-Vehicle Distance} \label{sec: 3-circle-intro} 
\begin{figure}[h]
\includegraphics[width=5cm]{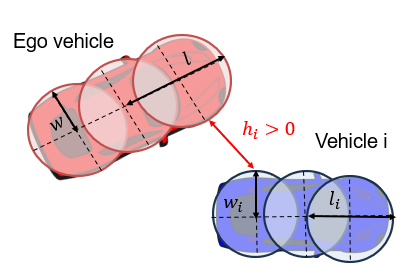}
\centering
\caption{Illustration of the inter-vehicle distance metric.}\label{fig:dist-metric}
\end{figure}

To integrate vehicle dimensions into our safety assessment, we model each vehicle using three circles. The metric for inter-vehicle distance, $h_i(x,y)$, between the ego vehicle (i.e., the vehicle to be controlled) and the $i$-th vehicle, is determined by evaluating the smallest distance between any pair of the evaluation points on the vehicle's body. Formally, this is given by 
\begin{equation}\label{eq:dist-metric}
    h_i(x,y) = \min_{p, q \in \{-1,0,1\}} d_i(p,q, x, y, \psi),
\end{equation}
\normalsize
where $
    d_i(p,q,x,y,\psi) = [((x + pD\cos\psi) - (x_i + qD_i\cos\psi_i))^2 
    + ((y + pD\sin\psi) - (y_i + qD_i\sin\psi_i))^2]^{\frac{1}{2}}
    - (w+w_i),
D = l-w, D_i = l_i-w_i$; $l$ and $w$ are the half-length and half-width of the ego vehicle, respectively, while $l_i$ and $w_i$ are the half-length and half-width of the $i$-th vehicle. This inter-vehicle distance metric is visualized in Fig.~\ref{fig:dist-metric}.

\begin{remark}
    To identify evaluation points on the vehicle using geometry according to~\eqref{eq:dist-metric},  we approximate the heading angle of ego vehicle $\psi$ and heading angle of $i$-th surrounding vehicle $\psi_i$  utilizing finite difference between two adjacent points on their state trajectories. 
\end{remark}
\begin{remark}
   Given the impracticality of directly obtaining the positions of other vehicles at every time step, we compute the distance measure $h_i(x,y)$ in~\eqref{eq:dist-metric} between the predicted positions of other vehicles and the ego vehicle trajectory.
\end{remark}
\subsubsection{Safety Specification for Collision Evaluation}
To ensure collision-free trajectories relative to other vehicles, we define the safety specification as
$
    h_i(x,y) \ge \epsilon,
$
where $\epsilon$ is a user-defined safety buffer. The safety check is executed between the ego vehicle and all surrounding vehicles. If any of these checks violate the safety specification, the collision penalty term in~\eqref{eq:fitness-all} is invoked.

\subsubsection{NN Constraints as Guidance for PSO} 
\begin{figure*}
        \centering
        \begin{subfigure}[b]{0.45\textwidth}
            \centering
            \includegraphics[width=\textwidth]{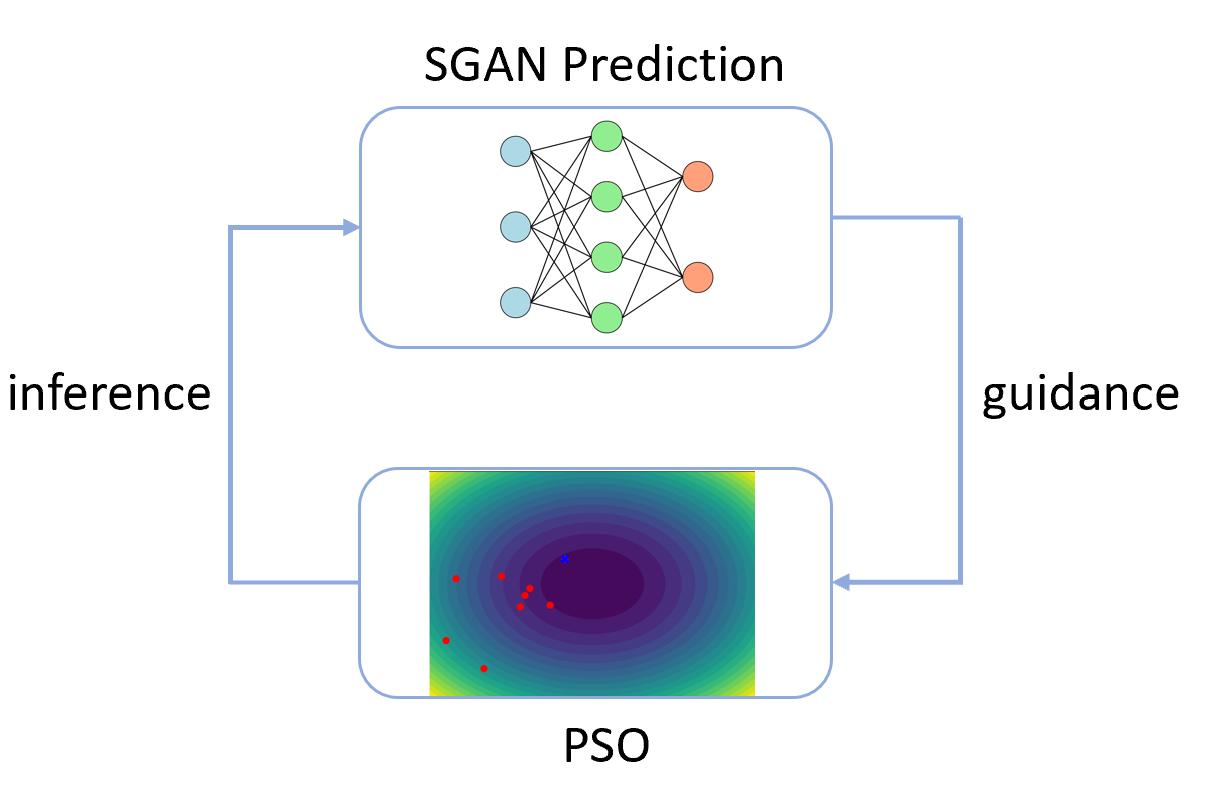}
            \caption{The architecture of PSO with SGAN prediction.}
        \end{subfigure}
        \hfill
        \begin{subfigure}[b]{0.4\textwidth}  
            \centering
        \begin{subfigure}[b]{0.4\textwidth}
            \centering
            \includegraphics[width=\textwidth]{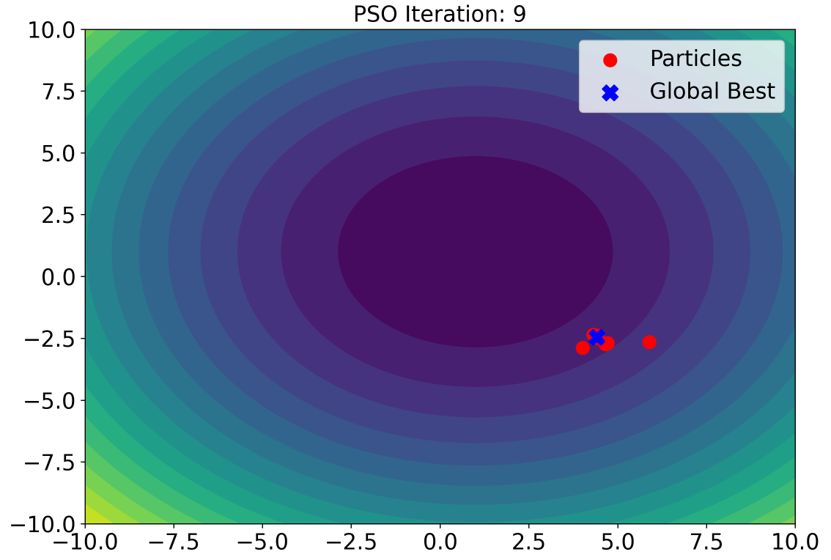}  
        \end{subfigure}
        \begin{subfigure}[b]{0.4\textwidth}  
            \centering 
            \includegraphics[width=\textwidth]{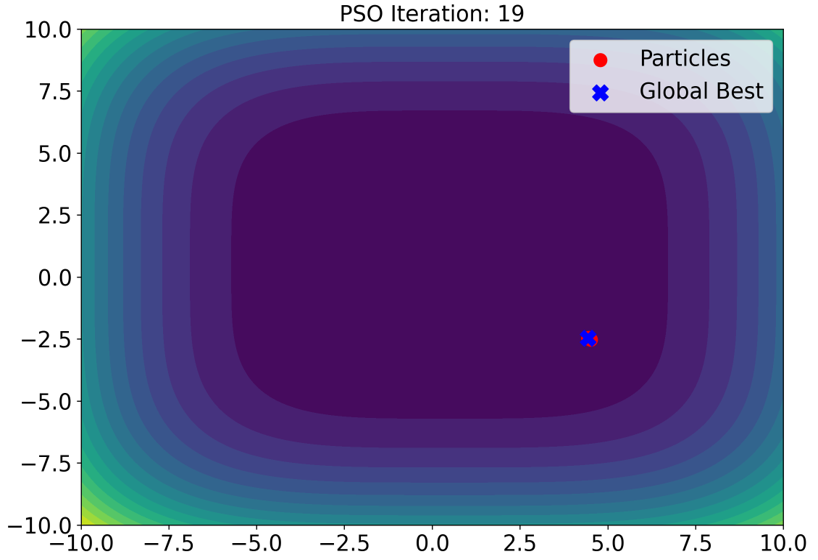}
        \end{subfigure}
        \vskip\baselineskip
        \begin{subfigure}[b]{0.4\textwidth}   
            \centering 
            \includegraphics[width=\textwidth]{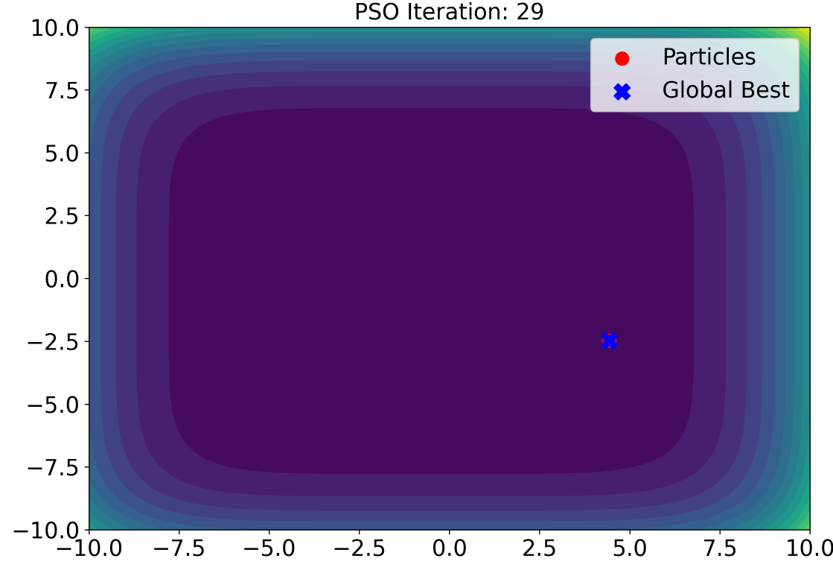}
        \end{subfigure}
        \begin{subfigure}[b]{0.4\textwidth}   
            \centering 
\includegraphics[width=\textwidth]{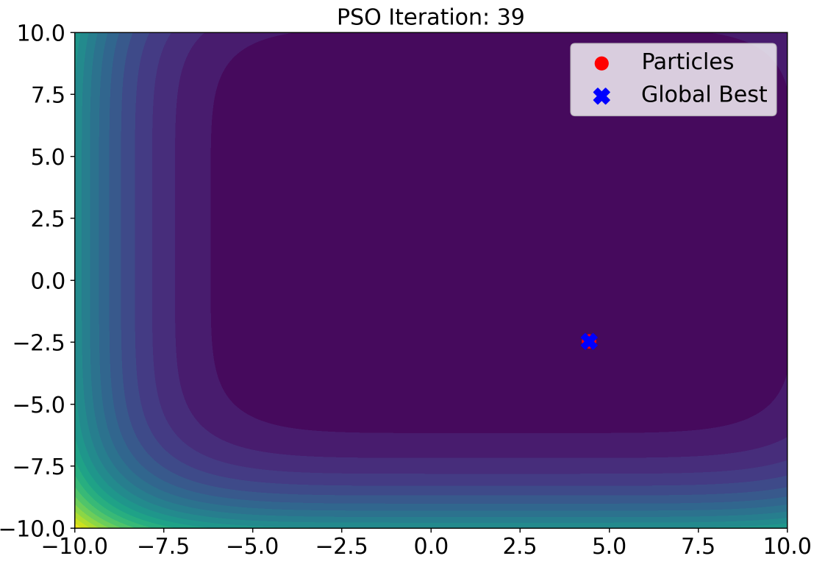}
        \end{subfigure}
        \caption{An illustrative example of particle updates and modified potential fields based on the NN outputs.}
        \end{subfigure}
        \caption
        {\small The overview architecture of PSO with the integrated SGAN prediction module (left) and an illustrative example of the evolution process of particles in PSO which are also updated based on the NN outputs (right). }
        \label{fig:pso_with_sgan}
    \end{figure*}

{Evaluating the interactive behavior of surrounding vehicles with respect to the ego vehicle's future action is crucial for a successful maneuver in dense traffic. This interaction is addressed by the integration of the neural network in the PSO search in a two-way manner. The overall architecture of incorporating SGAN prediction in the PSO search is presented in Figure~\ref{fig:pso_with_sgan}. The integrated NN component iteratively infers the future positions of other vehicles based on the particle position, which can be representative as the future position of the ego vehicle. On the other hand, the interactive behavior modeling capacities of SGAN are beneficial in updating the quality measure of particles and guiding the search process of PSO. The safety constraints imposed by the NN prediction modify the cost values of particles, and the particle positions are then updated in the modified potential field based on the updated predictions of surrounding vehicles. An illustrative example of the particle updates based on the modified cost values and potential fields based on the SGAN outputs is shown in Figure~\ref{fig:pso_with_sgan}. The reasoning engine of interactive behavior provided by SGAN facilitates the search for feasible and collision-free trajectories to achieve a successful maneuver in dense traffic.}

\subsection{Smooth Trajectory Generation from Key Waypoints}
Within the PSO iterations, a dynamically feasible trajectory is determined by propagating the best-found particles. This trajectory provides points $(\hat{x}_i,\hat{y}_i)$ that are utilized to construct a 3rd-order polynomial curve $\kappa(l)$, defined as $
    \kappa(l) = \beta_3 l^3 + \beta_2 l^2 + \beta_1 l + \beta_0$. 
 To orient the vehicle forward at $(x_f,y_f)$, the heading angle $\psi_f$ is set to 0, which is specified as the terminal condition of the fitted polynomial curve. 

The polynomial curve $\kappa(l)$ with waypoint enforcement preserves the trajectory smoothness. Subsequently, refined waypoints $(\bar{x}_i, \bar{y}_i, \bar{v}_i)$ are established by associating Cartesian coordinates from $\kappa(l)$ with vehicle velocity profiles using existing planners. These waypoints then serve as reference positions and guide the iterative PSO procedure.
A detailed depiction of the proposed trajectory planning algorithm is presented in Algorithm~\ref{alg:2}. 

\begin{algorithm}
    \caption{PSO-based trajectory planning for lane change}\label{alg:2}
    \begin{algorithmic}[1]
        \Require{Target state of the ego vehicle $(x^*,y^*,\psi^*)$, current state of the ego vehicle $(x_0, y_0, \psi_0)$, observed positions of other vehicles $[(z_{i,1},\dots, z_{i,N_{veh}})]_{i=1:N_{obs}}$}
        \Ensure{ Feasible trajectory for merging $[(\bar{x}_i, \bar{y}_i, \bar{v}_i)]_{i=1:N}$}
        \State Generate an initial trajectory $\bar{z}_{i0} = [(\bar{x}_i, \bar{y}_i, \bar{v}_i)]_{i=1:N}$ for the ego vehicle  using an existing planner 
        \State Predict positions of other vehicles $[(z_{i,1}, \dots, z_{i,N_{veh}})]_{i=1:N_{pred}}$ using the SGAN model introduced in~\ref{sec:SGAN-intro}
        \While{feasible trajectory is not found}
         \State Compute reference acceleration $\bar{a}_i$ and heading angle sequence $\bar{\psi}_i$ using reference waypoints $ [(\bar{x}_i, \bar{y}_i, \bar{v}_i)]_{i=1:N}$ 
         \State Use the PSO algorithm as described in Algorithm~\ref{alg:1} to find a dynamically feasible trajectory $[(\hat{x}_i,\hat{y}_i)]_{i=1:N}$ and a refined steering angle sequence $[\delta_1^*,\dots,\delta_N^*]$
        \State Fit a polynomial curve $\kappa(l)$ that passes through selected points from the feasible trajectory $[(\hat{x}_i,\hat{y}_i)]_{i=1:N}$ 
        \State Associate the fitted polynomial curve $\kappa(l)$ with reference waypoints using a planner and update the reference waypoints $[(\bar{x}_i, \bar{y}_i, \bar{v}_i)]_{i=1:N}$
        \EndWhile
    \end{algorithmic}
\end{algorithm}

\section{SIMULATION}
\subsection{Experimental Setups}
\subsubsection{Setup of Baseline Methods} 
\begin{figure}[h]
\includegraphics[width=5cm]{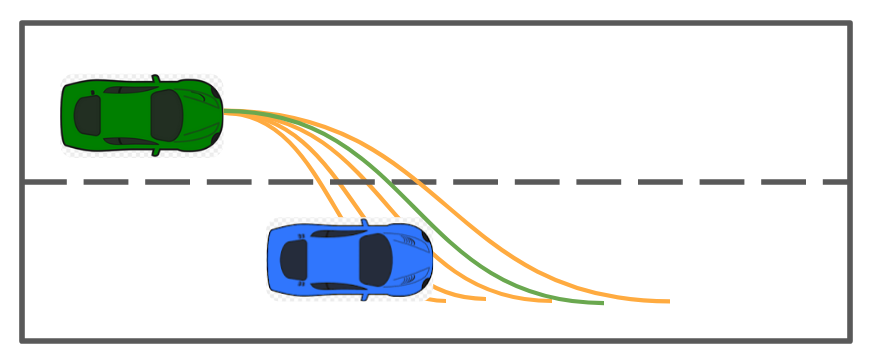}
\centering
\caption{Illustrative example of modifying planned trajectory using Monte-Carlo sampling.}\label{fig:modify-mc-intuit}
\end{figure}

We implement the trajectory planning technique utilizing the PSO algorithm for lane changes. For a comparative evaluation, our method is contrasted with ADMM-NNMPC~\cite{gupta2023interaction} --- an analytical approach that integrates neural networks into the MPC framework. 
Additionally, we explore a strategy that iteratively adjusts the target position of originally planned unsafe trajectories, ensuring post-modified trajectories satisfy safety constraints. We refer to this approach as \textit{Monte-Carlo (MC) sampling-based trajectory modification}. Figure~\ref{fig:modify-mc-intuit} provides a visual representation of this interaction-aware trajectory planning method based on target position adjustments. The green line in Figure~\ref{fig:modify-mc-intuit} represents a feasible trajectory that avoids all the other vehicles and maintains optimality for lane changes. 

\subsubsection{Experiment Platform} The experiments are conducted on a computer equipped with 11th Gen Intel Core i9-11900H (2.50GHz $\times$ 16 processors) and NVIDIA Corporation/Mesa Intel UHD graphics card, running Ubuntu 20.04 LTS. An unoptimized version of the proposed algorithm for trajectory planning leveraging PSO requires a computation time of less than 100 ms in most of the testing cases, showcasing the potential for real-time computation capability.
\subsubsection{Scenario Settings}\label{sec:experiment setting}We consider the trajectory planning for lane changes in dense traffic. Our goal is to achieve a smooth maneuver from the source lane to the target lane and avoid other vehicles. Optimized trajectories promote driving comfort and alignment with the lane center upon accomplishing the merge.

To validate the efficacy and effectiveness of our proposed planning approach, we set up a two-lane lane-changing experiment featuring three vehicles, including the ego vehicle and two other vehicles. Each vehicle measures 5m in length and 2m in width, while the lane's width is 3.5m. The initial positions of the other vehicles are randomized. We employ a trained SGAN model to predict other vehicles' behavior, thereby integrating safety constraints into the ego vehicle's trajectory planning as discussed in Section~\ref{sec:safety_constraints}. The SGAN's observation and prediction horizons are  8 and 12, respectively. 

\begin{figure}
\includegraphics[width=8.5cm]{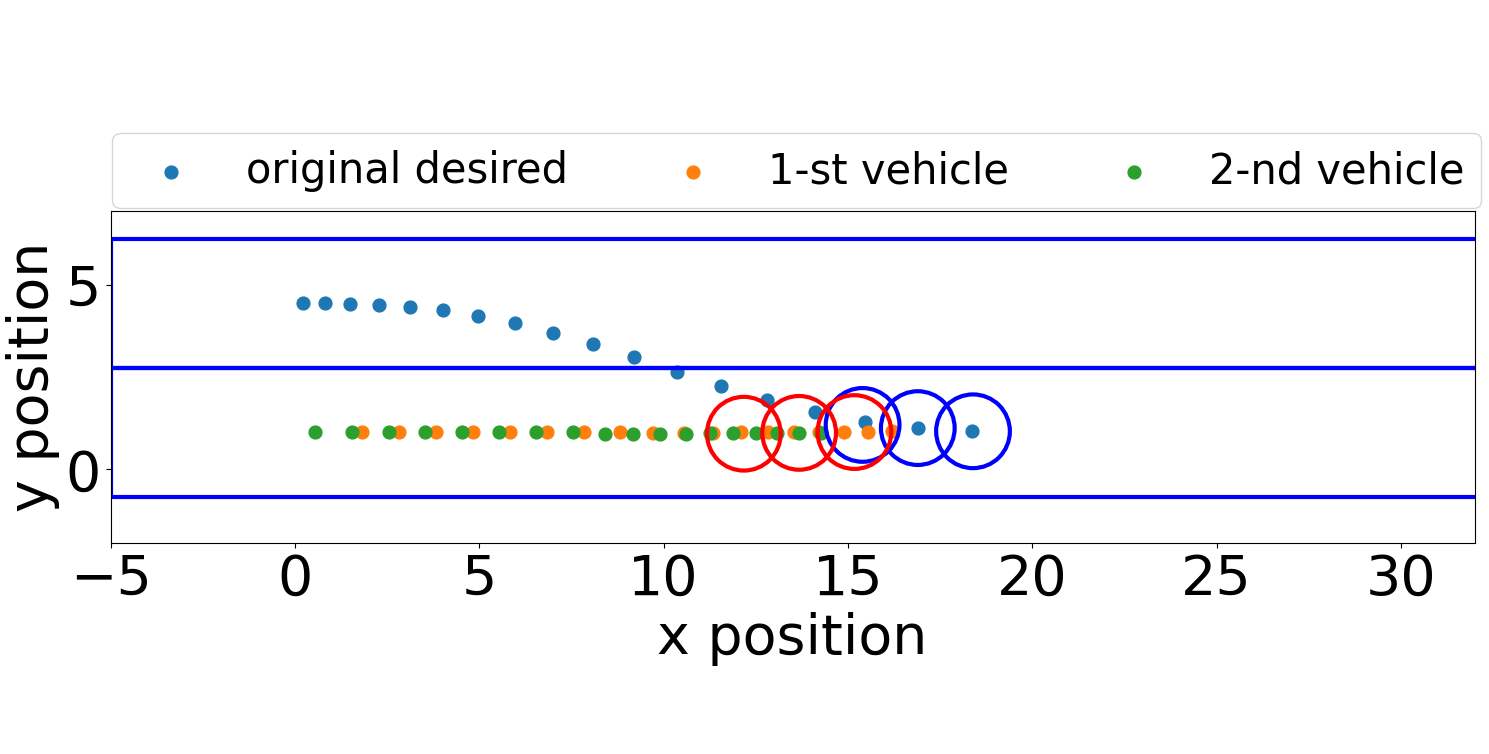}
\centering
\caption{Illustration of an initial trajectory plan that can cause a collision with other vehicles.
}\label{fig:init-plan}
\end{figure}

We consider the initial and desired positions of the ego vehicle to be well-aligned with the lane centers. The ego vehicle intends to merge to the bottom lane from the top lane as depicted in Figure~\ref{fig:modify-mc-intuit}.  The black lines in Figure~\ref{fig:modify-mc-intuit} and blue lines in Figure~\ref{fig:init-plan} represent the lane boundaries. The blue dots in Figure~\ref{fig:init-plan} denote the initial planned trajectory to merge into the target lane computed from an existing planner without considering other vehicles, and the initial trajectory plan can be unsafe. We use the three-circle representation introduced in Section~\ref{sec: 3-circle-intro} to represent the actual dimension of the vehicle, with each circle's radius being half of the vehicle width.  We identify the closest position between other vehicles and the ego vehicle during merging and highlight that position with three circles to represent the vehicles' actual configuration at the time step when the two vehicles are closest.  In Figure~\ref{fig:init-plan}, the blue circles represent the configuration of the ego vehicle when it gets closest to other vehicles, while the configuration of the closest other vehicle is denoted by the red circles. We observe the initial trajectory plan is unsafe and the ego vehicle can collide with other vehicles, shown by the overlap between the red and blue circles in Figure~\ref{fig:init-plan}.

For the ease of interpreting the vehicle trajectories, we produce Figures~\ref{fig:init-plan}-\ref{fig:modify-mc} for qualitative analysis. In the quantitative comparison results, we implement experiments in scenarios with more vehicles under small inter-vehicle gaps, which necessitates an interactive behavior reasoning module to achieve interaction-aware trajectory planning.

\subsection{Results}
\begin{figure}[hb]
\includegraphics[width=9cm]{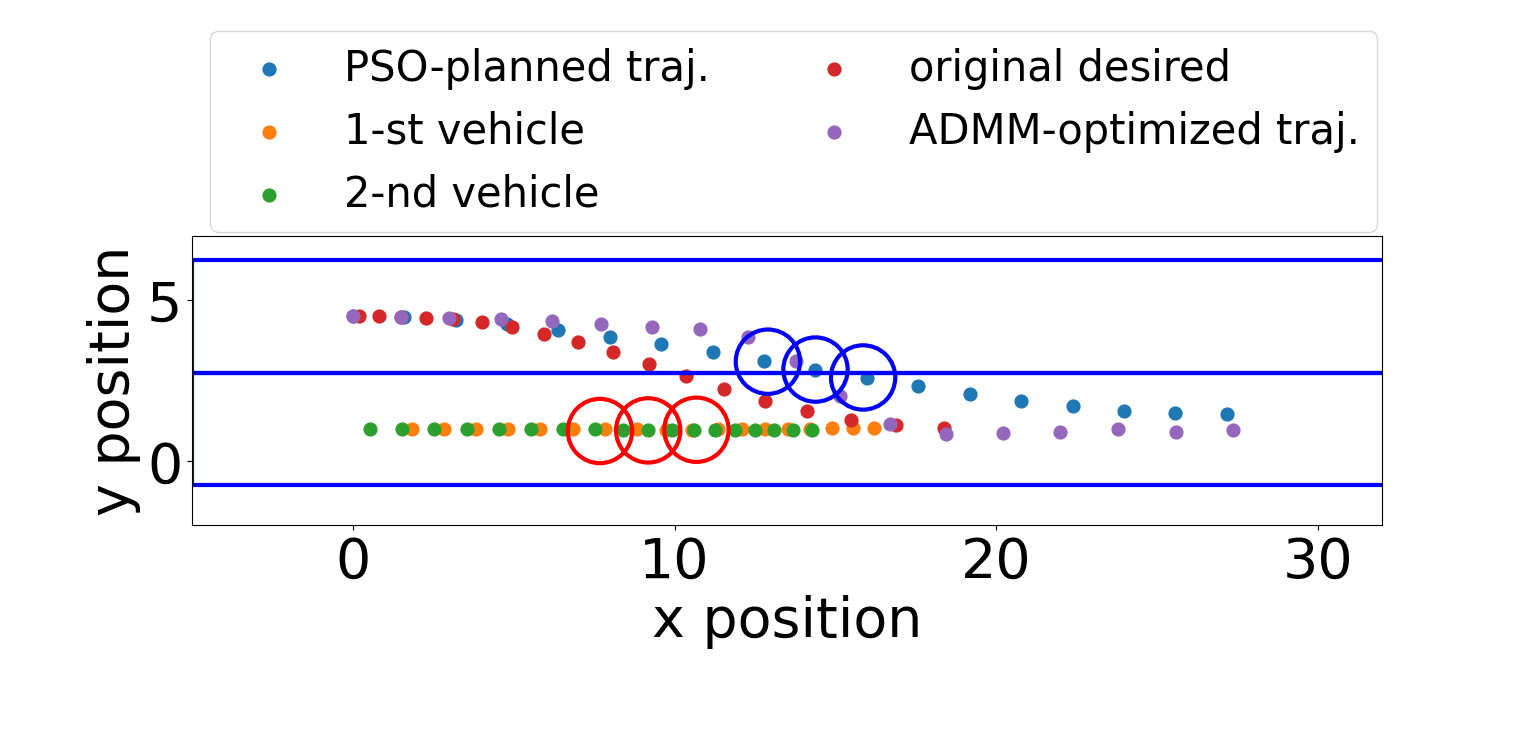}
\centering
\caption{Illustration of PSO-based trajectory planning approach and its comparison with ADMM-NNMPC.}\label{fig:comparison_admm_pso}
\end{figure}

 A key advantage of PSO-based planning is its potential for real-time computation.  In the scenario from Section~\ref{sec:experiment setting}, we implement the PSO-based planning algorithm for lane changes and conjecture that a feasible trajectory for safe lane changing can be computed within 100 ms in most testing scenarios. We choose the 100 ms computation time as a benchmark to evaluate the computation efficiency of our algorithm, considering it is sufficient for ROS communication with a 10-Hz frequency and can showcase the potential for hardware deployment on real vehicles. 
Figure~\ref{fig:comparison_admm_pso} contrasts the trajectory obtained from the PSO-based planning method with that from ADMM-NNMPC. The PSO-based planning method is capable of refining the initial unsafe trajectory and offering optimality comparable to ADMM-NNMPC. Notably, the PSO-based planning method requires less computation time and exhibits real-time computation capability.  
 
To validate the proposed planning method in more general scenarios, we randomly initialize the horizontal positions of other vehicles and implement 50 tests. Potential collisions between other vehicles resulting from the position random initialization are not factored into evaluating the performance of the planning algorithm. Here, we interpret a feasible solution determined within 0.2s of computation time by the planning algorithm as a `success' and record the success rate among the 50 experiments. We highlight that computation efficiency is also an important metric to evaluate the success rate of a planning algorithm since an accurate solution can be impractical for real-time deployment.

\begin{figure}[!h]
\includegraphics[width=9cm]{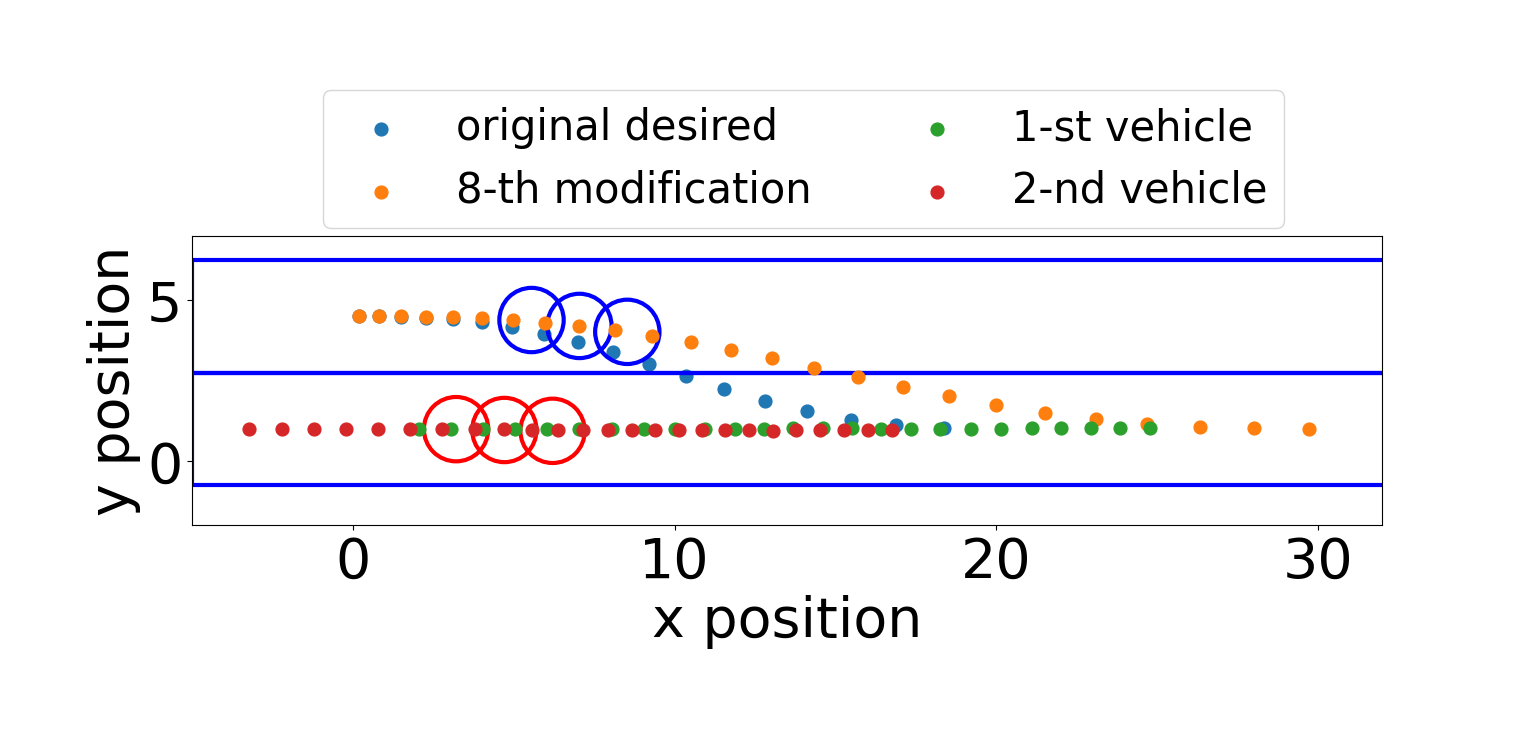}
\centering
\caption{Illustration of modifying original infeasible trajectories based on Monte-Carlo sampling.}\label{fig:modify-mc}
\end{figure}

In the MC-sampling-based trajectory modification method, the success rate indicates the percentage of experiments that a feasible trajectory can be obtained among 12 modifications within a  margin of 15m. An illustration of a feasible trajectory obtained by modifying the original unsafe trajectory is presented in Figure~\ref{fig:modify-mc}, in which a trajectory that can avoid other vehicles is determined in the 8th modification. Different methods to address the trajectory planning problem are compared in the aspects of success rate, minimum distance to other vehicles during merging, time steps needed to complete the merge, and the average computation time. The comparison results are summarized in Table~\ref{table_example}.

\begin{table}[h]
\caption{Comparison with other methods}
\label{table_example}
\begin{center}
\begin{tabular}{|c|c|c|c|c|}
\hline
method & \shortstack{success \\rate (\%)} & \shortstack{min dist to \\other veh. (m)} & \shortstack{avg. comp. \\time (ms)} & \shortstack{\# of step \\to merge}\\
\hline
PSO (2-particle) & \textbf{85} & 2.114 & 176 & 18\\
\hline
\shortstack{ADMM (\textbf{Oracle})} & 100 & -- & 22264 & \textbf{10}\\
\hline
MC  & 78 & 2.016 & \textbf{125.1} & 20-25\\
\hline
\end{tabular}
\end{center}
\end{table}

From Table~\ref{table_example}, all three methods ensure a safe merge into the target lane, maintaining a minimum distance to other vehicles over 2m (safety buffer) throughout the horizon. The PSO-based planning approach achieves a higher success rate than the MC-based trajectory modification. In terms of trajectory optimality, the PSO-based approach outperforms the latter, achieving trajectory planning with fewer time steps to complete the merge. Though the ADMM-NNMPC method delivers the most accurate solution and superior lane change trajectory regarding the number of required time steps, the result is accomplished at the cost of computation efficiency.

\begin{table}[h]
\caption{Comparison between various particle numbers in the PSO-based trajectory planning}
\label{table_example_2}
\begin{center}
\begin{tabular}{|c|c|c|c|c|c|}
\hline
particle no. & 1 & 2 & 3 & 4 & 5\\
\hline
avg. comp. time (ms) & 411 & \textbf{176} & 289 & 308 & 373\\
\hline
success rate (\%) & 20 & 85 & \textbf{90} & 55 & 85\\
\hline
\end{tabular}
\end{center}
\end{table}

{We perform experiments to evaluate the efficacy of various swarm sizes (number of candidate solutions), and the results are outlined in Table~\ref{table_example_2}. For the trajectory planning problem, formulating the PSO with 2 or 3 particles provides the best overall performance in terms of both success rate and computation time. In our present algorithm, as well as in the standard PSO method, particles' positions and velocities are updated sequentially. Compared with the 1-particle setup, using a 2 or 3-particle configuration in PSO enhances the information exchange within the swarm, which facilitates the search for the global best particle and even offsets the additional computation involved in the sequential particle updates. Nevertheless, as the number of particles further increases, the computational cost becomes more significant, outweighing the advantages of expedited global-best particle search provided by the information exchange between particles. Considering the algorithm success rate is computation-efficiency dependent and any unfinished computation within 0.2s computation time is interpreted as a failure, we observe the 5-particle setup in PSO fails to further increase the success rate due to the increased computation burden and the average computation time also increases. We also observe setups with more particles are impractical regarding improving the algorithm success rate and the potential for hardware deployment. However, incorporating an appropriate parallel computation design for particle updates can reduce the computation time and further improve the algorithm optimality, which we defer as future work topics. }

\begin{remark}
    In simulation scenarios where the initial planned trajectory is already feasible and ensures collision-free merging, the PSO-based planning minimally deviates from the initial planned trajectory to retain optimality.
\end{remark} 


\section{CONCLUSION}
In this paper, we address computation-efficient, interaction-aware trajectory planning in dense traffic. We leverage the social generative adversarial network (SGAN) model to predict surrounding vehicles' behavior, ensuring safety by imposing constraints on trajectory planning. Particle swarm optimization aids in efficiently computing a collision-free lane-change trajectory, with candidate solutions updated based on neural network predictions. The proposed trajectory planning algorithm exhibits real-time computation capability. Numerical examples from a two-lane scenario validate its effectiveness and efficiency, with comparative results demonstrating its superiority or parity of planned trajectory in the aspects of optimality and success rate. Notably, the proposed planning algorithm enhances the computation efficiency. Future research will explore uncertainty quantification in vehicle behavior predictions and mitigation strategies through planning and control techniques.

\bibliography{reference}
\bibliographystyle{unsrt}


\end{document}